\documentclass{article}
\usepackage{spconf,amsmath,graphicx}
\usepackage{cite}
\usepackage{amssymb,amsfonts}
\usepackage{algorithmic}
\usepackage{textcomp}
\usepackage{xcolor}
\usepackage{subcaption}
\usepackage{adjustbox}
\usepackage{caption}

\title{Unlocking Deep Learning: A BP-Free Approach for Parallel Block-Wise Training of Neural Networks}
\name{
    \begin{tabular}[t]{c}
        Anzhe Cheng$^{*}$ \qquad Zhenkun Wang$^{*}$ \qquad Chenzhong Yin$^{*}$ \qquad Mingxi Cheng$^{*}$ \\ \qquad Heng Ping\sthanks{All authors contributed equally to this work} \qquad Xiongye Xiao \qquad Shahin Nazarian \qquad Paul Bogdan
    \end{tabular}
}

\address{University of Southern California, Los Angeles, CA, USA }

\begin{document}
%
\maketitle
\begin{abstract}
Backpropagation (BP) has been a successful optimization technique for deep learning models. However, its limitations, such as backward- and update-locking, and its biological implausibility, hinder the concurrent updating of layers and do not mimic the local learning processes observed in the human brain. To address these issues, recent research has suggested using local error signals to asynchronously train network blocks. However, this approach often involves extensive trial-and-error iterations to determine the best configuration for local training. This includes decisions on how to decouple network blocks and which auxiliary networks to use for each block.
In our work, we introduce a novel BP-free approach: a block-wise BP-free (BWBPF) neural network that leverages local error signals to optimize distinct sub-neural networks separately, where the global loss is only responsible for updating the output layer. The local error signals used in the BP-free model can be computed in parallel, enabling a potential speed-up in the weight update process through parallel implementation. Our experimental results consistently show that this approach can identify transferable decoupled architectures for VGG and ResNet variations, outperforming models trained with end-to-end backpropagation and other state-of-the-art block-wise learning techniques on datasets such as CIFAR-10 and Tiny-ImageNet. The code is released at https://github.com/Belis0811/BWBPF. 
\end{abstract}
\begin{keywords}
Local loss, Block-wise learning, Computer vision
\end{keywords}
\vspace{-5mm}
\section{Introduction}
\vspace{-1mm}
\label{sec:intro}

Based on the remarkable success of gradient descent training techniques~\cite{pruthi2020estimating}, the Backpropagation (BP) algorithm~\cite{rumelhart1986learning} has emerged as a widely adopted core learning algorithm in most deep learning networks. This algorithm efficiently calculates weight parameter gradients by reverse-propagating the error signal from the loss function to each layer. However, BP faces certain limitations, characterized by update locking and backward locking~\cite{jaderberg2017decoupled}, where update locking necessitates the completion of a forward pass before any weight updates can occur, while backward locking requires gradient computation in upper layers to precede that in lower layers.

An intriguing category of alternative proposals is grounded in contrastive learning within energy-based models~\cite{xie2003equivalence, bengio2015early, scellier2017equilibrium}. Rather than propagating errors layer by layer through random feedback connections, Nøkland \cite{nokland2016direct} and Neftci \textit{et al.}'s networks \cite{neftci2017event} employ a fixed random projection of the top layer error as the error signal in deep layers, which enables a unified error signal shared across all layers. However, this approach still leads to substantial delays and memory demands during weight updates. This occurs because a full forward pass through the entire network must be completed before the error signal becomes accessible, requiring deep layers to maintain their states throughout the entire forward pass, which consumes significant computational resources.


\begin{figure*}[htbp]
 \centering
    \includegraphics[width=\textwidth]{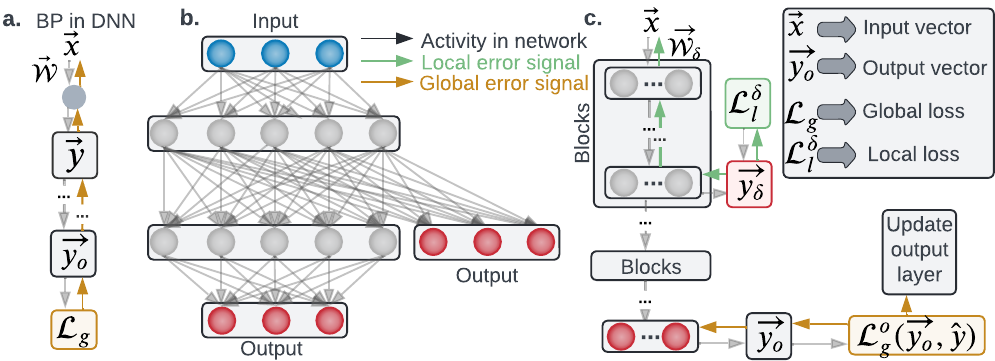}
\caption{\textbf{Weight updates of BWBPF.} \textbf{a.} BP in classic neural network training. Global prediction loss is back-propagated through layers. \textbf{b.} The dense layer is attached below the subnetwork for computing the local loss. \textbf{c.} BP-free weight update of a block of layers.}
\label{fig1}
\end{figure*}

To address these challenges, researchers have introduced a solution in the form of greedy block-wise learning approaches~\cite{belilovsky2020decoupled,lowe2019putting}. These methods involve partitioning the entire network architecture into separate, distinct subnetworks or blocks.
Each block consists of one or more consecutive layers and is then connected to a compact neural network referred to as an auxiliary network. This auxiliary network computes the local loss and employs it to optimize the weights of the layers within the block by propagating error signals. Crucially, this scheme enables individual blocks to independently execute parameter updates, even as other blocks are concurrently engaged in forward pass computations. Recently, 
Gunhee \textit{et al.}~\cite{pyeon2020sedona} proposed an automatic searching method named SEDONA, which facilitates the efficient exploration of decoupled neural architectures in the context of greedy block-wise learning. Starting with a base neural network, SEDONA enhances validation loss optimization by organizing layers into blocks and determining the most suitable auxiliary network for each block. However, when using deeper networks with a larger number of blocks, SEDONA's classification performance either shows only marginal improvement or remains similar to that of the standard BP method.

In this paper, to fully utilize the potential of the block-wise local loss approach, we present a novel BP-free learning approach based on the block-wise BP-free (BWBPF) learning method. 
Leveraging the original architecture's blocks as a foundation, we partition the network into several subnetworks connected via our auxiliary network. 
These subnetworks are connected to independent dense layers, which are responsible for calculating local losses. These local loss terms are then utilized to update the parameters of each individual block.
Through experiments on two image classification datasets, CIFAR-10 and Tiny-ImageNet, our method outperforms the original BP algorithm in ResNet\cite{he2016deep} and VGG\cite{simonyan2014very} networks. 
Furthermore, we compare our results with three greedy learning algorithms, DGL~\cite{belilovsky2020decoupled}, PredSim~\cite{nokland2019training}, and the SEDONA method~\cite{pyeon2020sedona}. Our network consistently outperforms networks trained using those algorithms. Notably, unlike most of the aforementioned methods, our approach doesn't necessitate additional memory blocks to generate an error signal.

\vspace{-4mm}
\section{Method}
\vspace{-1mm}
\label{sec:method}

The BWBPF learning approach draws inspiration from biological networks such as the human brain, where synaptic weight updates can occur through local learning, independent of the activities of neurons in other brain regions~\cite{caporale2008spike, yin2023anatomically, imms2023early}. Partly for this reason, local learning has been identified as an effective means to reduce memory usage during training and to facilitate parallelism in deep learning architectures. 

\subsection{Block-wising learning in convolutional blocks}

In the context of a convolutional neural network (CNN) comprised of multiple layers, the concept of "blocks" emerges as a convenient way to organize and group one or more convolutional layers together. To simplify this concept further, when dealing with a neural network featuring multiple convolutional layers, we can naturally group a set of these layers together into what we refer to as a "block." This grouping approach is widely employed in well-known architectures like VGG~\cite{simonyan2014very}, ResNet~\cite{he2016deep}, and Inception~\cite{szegedy2015going}, where these blocks play a pivotal role in shaping the network's structure and behavior.

Following each convolutional block in our model, we incorporate two essential components: a global average pooling layer and a dense layer responsible for computing local loss. In this work, we exclusively employ cross-entropy loss functions. These auxiliary components within the network, as depicted in Fig.~\ref{fig1} (b), hold the crucial responsibility of updating the layer weights within each block. Importantly, they carry out this weight-updating task while preserving the overall structural integrity of the network, ensuring that the underlying architecture remains intact. This architectural choice enables our network to consistently deliver superior performance when compared to networks trained using alternative algorithms. It is noteworthy that our approach achieves this performance without the need for additional memory blocks to generate an error signal, a departure from many of the existing methodologies in the field.

\vspace{-4mm}
\subsection{BP-free}
\vspace{-2mm}

To overcome the drawbacks of BP, particularly the issue of backward locking, we propose the BWBPF learning algorithm which eliminates BP for the global prediction loss and instead computes the local prediction loss. This modification implies that the global prediction loss is solely responsible for optimizing the output layer, denoted as $\mathcal{L}_g^o$ (where $o$ stands for output). In simpler terms, this prediction loss acts as a local loss function exclusively for updating the weights of the output layer (as depicted in Fig.~\ref{fig1} (c)).
The total loss function for the BP-free algorithm is defined as follows:
\vspace{-1mm}
\begin{equation}
\label{eq:bpfree}
    \mathcal{L} = \lambda_1\mathcal{L}_g^o + \lambda_2\sum_{l=1}^{K}\mathcal{L}_l^{\delta}
\end{equation}
where $\mathcal{L}_l^{\delta}$ represents the local loss, with $l$ denoting the number of blocks. The local loss is computed between the true label and the local output vector ($\vec{y}_\delta$). By removing the BP of the global prediction loss to the hidden layers, the weight updates for the convolutional layers in our model are exclusively influenced by the local prediction loss, as illustrated in Figure \ref{fig1} (c).

\begin{figure}[htbp]
 \centering
 \begin{subfigure}{0.49\textwidth}
    \centering\includegraphics[width=1.05\linewidth]{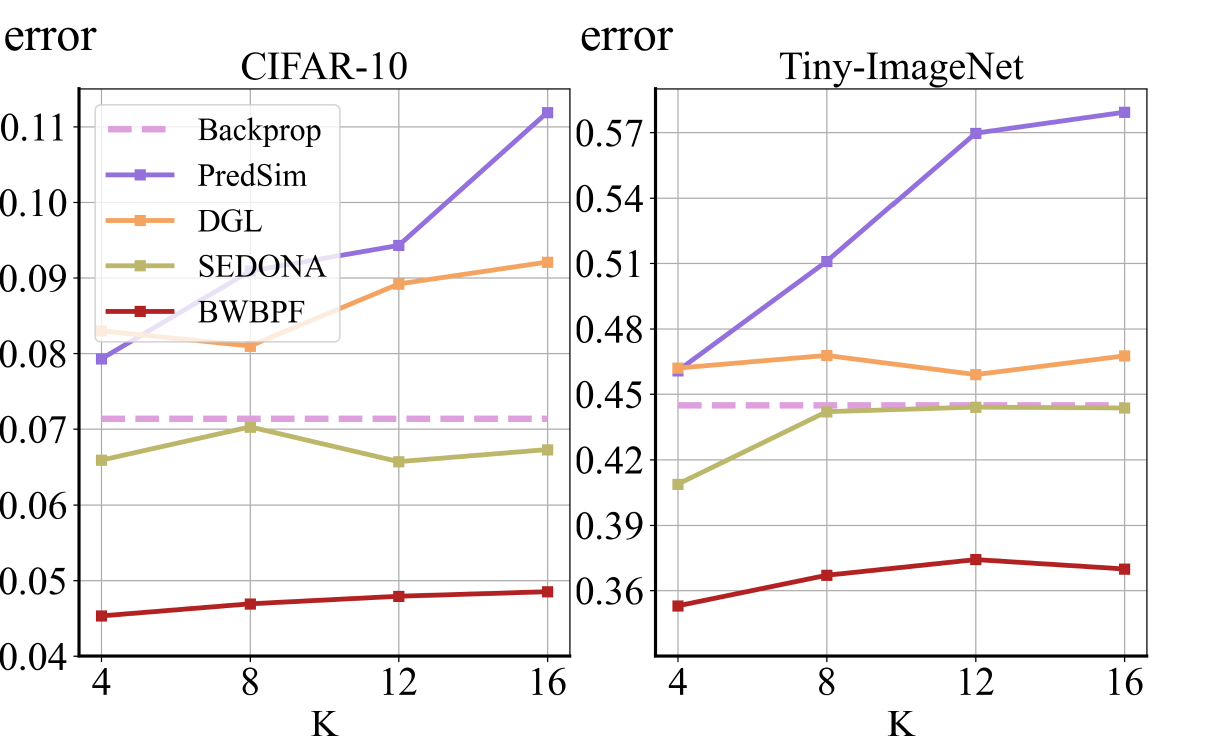}
    \caption{ResNet101}
  \end{subfigure}
  \begin{subfigure}{0.49\textwidth}
    \centering\includegraphics[width=1.05\linewidth]{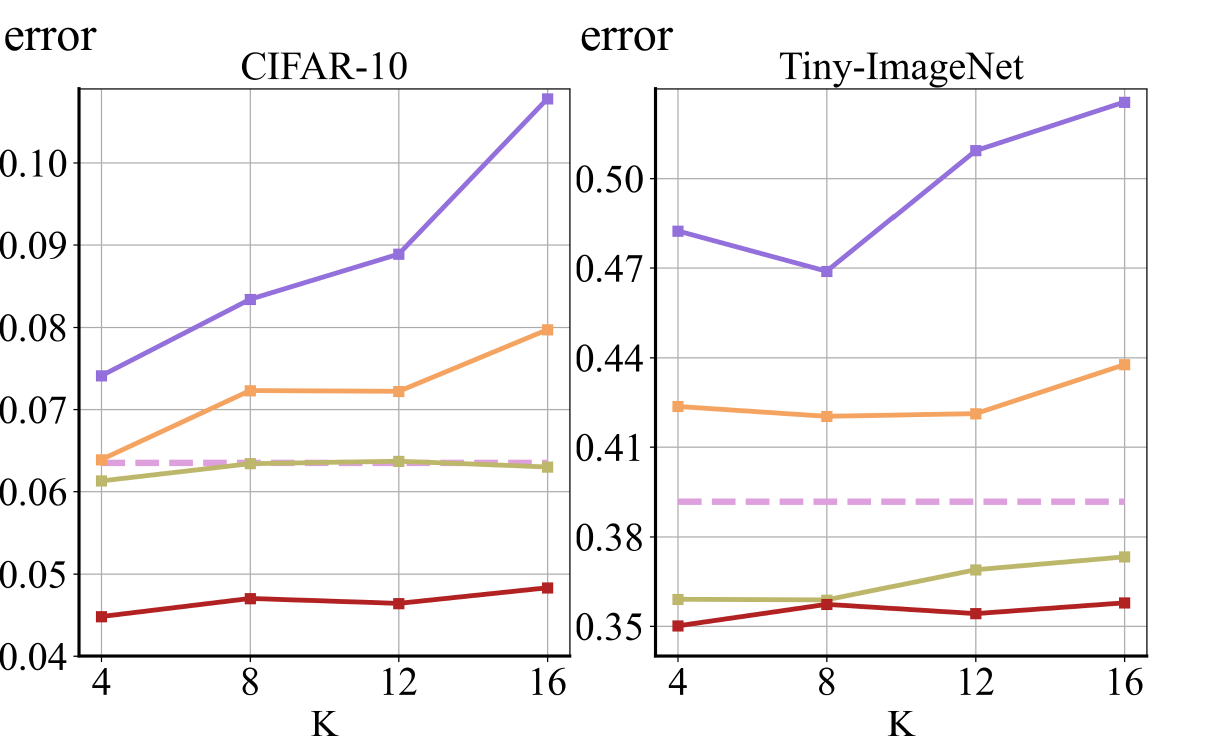}
    \caption{ResNet152}
  \end{subfigure}
\caption{Classification error (1-accuracy) curves of (a) ResNet-101 and (b) -152 with increasing \textit{K} on CIFAR-10 and Tiny-ImageNet. }
\label{fig2}
\end{figure}

To enable layers in convolutional blocks to generate valid predictions, we apply the same activation function used in the output layer for each block. For instance, in the case of a CNN designed for $N$-class classification, we typically include $N$ output neurons in the output layer and apply the softmax function. 
In this work, we equally group the layers into $K$ different blocks and embed our approach on VGG-19, ResNet50, ResNet101 and ResNet152 for validating our model's performance by spilt the original architecture into 4, 8, 12, and 16 blocks. 
We conducted model training for 200 epochs with CIFAR-10 and 300 epochs with Tiny-ImageNet, using a batch size of 32. We initially set the learning rate to 0.1 and gradually reduced it to 0.0001. For the ResNet-50/101/152 and VGG-19 architectures, we applied a weight decay of 0.0001 and a momentum of 0.9. Additionally, we utilized weight initialization without employing dropout.

\vspace{-1mm}
\section{Experiments}
\label{sec:experiment}
\vspace{-1mm}

To validate our approach and make a meaningful comparison with state-of-the-art block-wise learning methods, we employ our approach and baseline methods on both VGG-19\cite{simonyan2014very} and ResNet-50/101/152\cite{he2016deep} architectures. 
To ensure a fair comparison, all methods are trained and tested on identical datasets, which include CIFAR-10 and Tiny ImageNet. Furthermore, we maintain uniformity in the training strategy and hyperparameter settings across all approaches.

\begin{table}[h!]
 \small
 \caption{Error rates(\%) on CIFAR-10 (a) and Tiny-ImageNet (b) with different methods with 4 outputs}
 \vspace{-1mm}
 \begin{subtable}{0.48\textwidth} 
 \begin{adjustbox}{width=\columnwidth,center}
   \begin{tabular}{cccccc}
   \hline
   \textbf{Architecture}&$^{\mathrm{*}}$\textbf{BP}&$^{\mathrm{*}}$\textbf{PredSim}&$^{\mathrm{*}}$\textbf{DGL}&$^{\mathrm{*}}$\textbf{SEDONA}&\textbf{BWBPF} \\
   \hline
   VGG-19 & 12.31 & 13.87 & 12.19 & 11.58 & \textbf{6.52}\\
   ResNet50 & 7.99 & 8.93 & 8.27 & 7.53 & \textbf{4.85}\\
   ResNet101 & 7.14 & 7.93 & 8.30 & 6.59 & \textbf{4.53}\\
   ResNet152 & 6.35 & 7.41 & 6.39 & 6.13 & \textbf{4.48}\\
   \hline
   \multicolumn{6}{l}{$^{\mathrm{*}}$Results are from SEDONA.} 
   \end{tabular}
   \end{adjustbox}
   \caption{CIFAR-10}
   \label{tab:subtaba}
 \end{subtable}

 \begin{subtable}{0.48\textwidth} 
  \begin{adjustbox}{width=\columnwidth,center}
   \begin{tabular}{cccccc}
   \hline
   \textbf{Architecture}&$^{\mathrm{*}}$\textbf{BP}&$^{\mathrm{*}}$\textbf{PredSim}&$^{\mathrm{*}}$\textbf{DGL}&$^{\mathrm{*}}$\textbf{SEDONA}&\textbf{BWBPF} \\
   \hline
   VGG-19 & 47.11 & 55.30 & 48.70 & 43.44 & \textbf{40.09}\\
   ResNet50 & 46.54 & 55.22 & 46.04 & 45.60 & \textbf{35.83}\\
   ResNet101 & 44.50 & 46.80 & 46.20 & 40.88 & \textbf{35.28}\\
   ResNet152 & 39.18 & 48.24 & 42.36 & 35.90 & \textbf{35.01}\\
   \hline
   \multicolumn{6}{l}{$^{\mathrm{*}}$Results are from SEDONA.}
   \end{tabular}
    \end{adjustbox}
   \caption{Tiny-ImageNet}
   \label{tab:subtabb}
 \end{subtable}
 \label{tab1}
 \vspace{-3em}
\end{table}

\vspace{-2mm}
\subsection{Datasets}
\vspace{-1mm}
The CIFAR-10 is obtained from the TensorFlow datasets \cite{abadi2016tensorflow}. CIFAR-10~\cite{krizhevsky2009learning} consists of 60,000 images, each of size $32\times32$.
Tiny ImageNet~\cite{le2015tiny} consists of a dataset of $100,000$ images distributed across 200 classes, with 500 images per class for training, and an additional set of $10,000$ images for testing. All images in the dataset are resized to $64\times64$ pixels.

\vspace{-3mm}
\subsection{Baselines models}
\vspace{-1mm}
In this paper, we choose three state-of-the-art block-wise learning algorithms which use BP-free optimization, which are 
DGL\cite{belilovsky2020decoupled}, PredSim\cite{nokland2019training}, and SEDONA\cite{pyeon2020sedona}: (1) DGL uses greedy block-wise learning method to train their network by dividing it into K subnetworks. The auxiliary network they use denoted as MLP-SR-aux, which consists of one pooling layer with three point-wise convolutions followed by an average pooling layer and a 3-layer MLP; (2) PredSim combines two local losses, similarity matching loss and cross-entropy loss, which require two different auxiliary networks, an average pooling layer with fully connected layer and a convolutional layer, respectively; 
(3) SEDONA generates four auxiliary candidates, encompassing a sequence of components: a point-wise convolutional layer, a depth-wise convolutional layer, a specific number of inverted residual blocks, and ultimately concluding with a point-wise convolutional layer. Subsequently, each candidate undergoes further processing, involving the application of an average pooling layer and the integration of a fully connected layer.

\begin{figure}[htbp]
 \centering
 \begin{subfigure}{0.49\textwidth}
    \centering\includegraphics[width=1.05\linewidth]{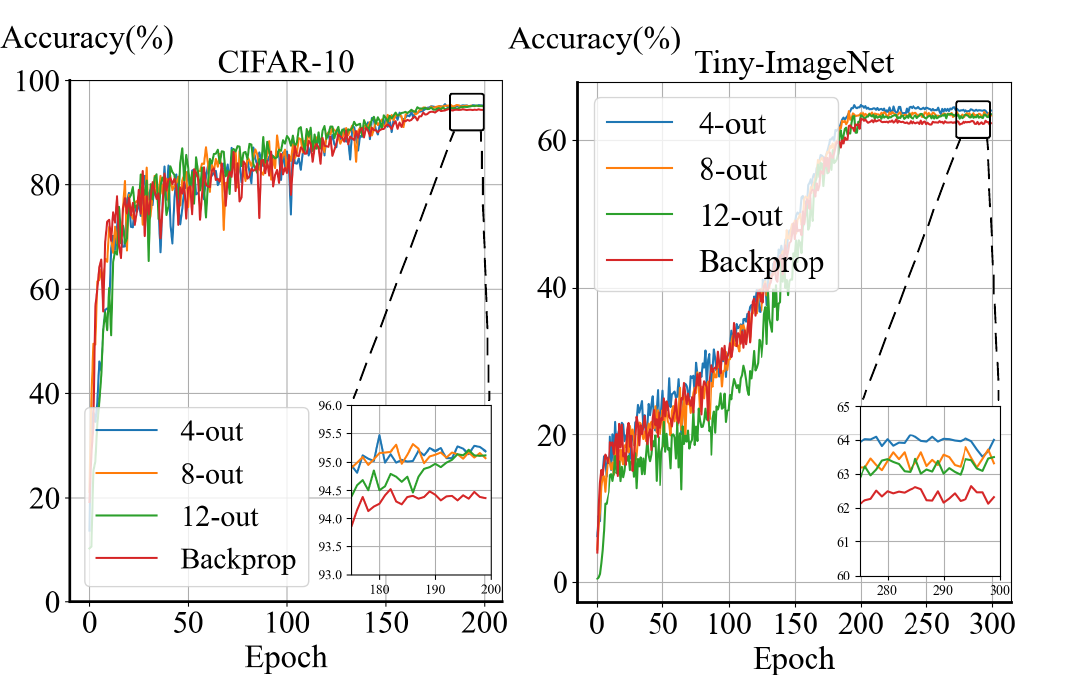}
    \caption{ResNet101}
  \end{subfigure}
  \begin{subfigure}{0.49\textwidth}
    \centering\includegraphics[width=1.05\linewidth]{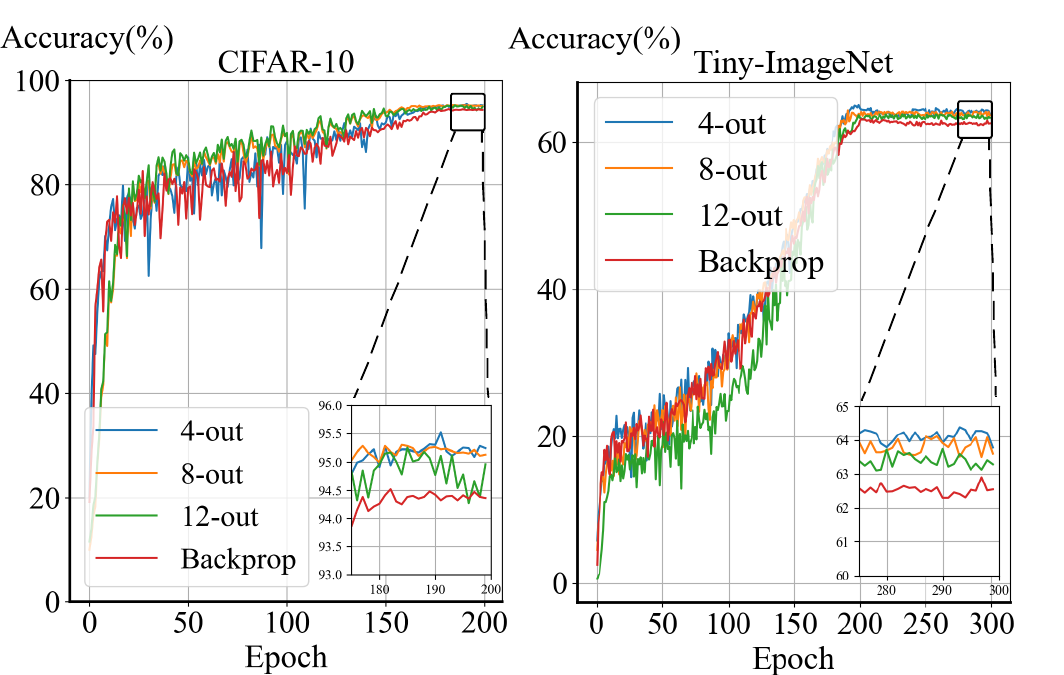}
    \caption{ResNet152}
  \end{subfigure}
\caption{Classification accuracy curves of (a) ResNet-101 and (b) -152 with increasing \textit{K} and end-to-end BP on CIFAR-10 and Tiny-ImageNet. }
\label{fig3}
\end{figure}

\vspace{-8mm}
\subsection{Results on CIFAR-10 and Tiny-ImageNet}

Our primary classification error rates (1-accuracy) for both CIFAR-10 and Tiny-ImageNet datasets are summarized in Table~\ref{tab1} (BP results are also shown in Table~\ref{tab1} for reference). For the CIFAR-10 dataset, noted that two state-of-the-art block-wise learning methods, PredSim and DGL, almost underperform traditional BP in all cases. In contrast, the network trained with the SEDONA method shows a better performance than BP. However, our model has a more significant improvement than the SEDONA method. Our image classification accuracy is about \textbf{5\%} higher on VGG-19 than on SEDONA, and about \textbf{2\%} higher on all three ResNet networks. 

Similarly, our method not only outperforms BP but also yields better performances to SEDONA on Tiny-ImageNet. As indicated in Table~\ref{tab:subtabb}, our approach's accuracy even surpassed SEDONA by over \textbf{10\%} on ResNet50. For the remaining ResNet architectures, the optimization gain from our method gradually diminished as the model complexity increased with additional layers. Nevertheless, our method consistently outperforms SEDONA across all configurations. 
We achieved a \textbf{5\%} improvement in accuracy compared to SEDONA with ResNet101, and we also observed a slight performance boost over SEDONA with ResNet152.
These results affirm that our method presents a superior option for both large and small datasets compared to conventional BP and other block-wise learning models.

To investigate the impact of the number of blocks ($K$) on model performance, Figure~\ref{fig2} presents results from experiments conducted using four different learning approaches: DGL, PredSim, SEDONA, and our BWBPF learning, each with varying values of $K$.
Upon analyzing the baseline methods, it is evident that both DGL and PredSim consistently underperform compared to the standard BP method across all values of $K$, particularly for $K \geq 12$. On the other hand, SEDONA exhibits superior performance to BP, particularly for lower values of $K$. However, as $K$ increases, SEDONA's performance becomes comparable to that of BP, especially when validating on CIFAR-10 and Tiny-ImageNet datasets using the ResNet 101 model.
In our model, we observe that increasing the value of $K$ leads to an increase in the error rate. Nevertheless, our model consistently outperforms the classic BP algorithm and all the baseline models in terms of overall model performance, even as $K$ increases.

\vspace{-3mm}
\section{Conclusion}
\label{sec:Con}
\vspace{-2mm}
To address the challenges posed by standard BP, we introduce a novel approach called the BWBPF learning algorithm, which operates without relying on BP. 
BWBPF, which divides the network into subnetworks connected to auxiliary networks, enables the computation of local losses for individual blocks, resulting in enhanced convergence, reduced gradient issues, better generalization, and increased model robustness. 
Through comprehensive experiments on image classification tasks using popular architectures such as ResNet and VGG, we showcase the superiority of our approach over traditional BP and existing state-of-the-art greedy learning methods. BWBPF's adaptability to various network architectures and datasets indicates its potential for broader applications in the field of deep learning. As we continue to explore alternative learning mechanisms inspired by biology, the proposed approach offers a promising avenue for improving the efficiency and effectiveness of training CNN architectures.


\bibliographystyle{IEEEbib}
\bibliography{strings,refs}

\begin{thebibliography}{10}

\bibitem{pruthi2020estimating}
Garima Pruthi, Frederick Liu, Satyen Kale, and Mukund Sundararajan,
\newblock ``Estimating training data influence by tracing gradient descent,''
\newblock {\em Advances in Neural Information Processing Systems}, vol. 33, pp.
  19920--19930, 2020.

\bibitem{rumelhart1986learning}
David~E Rumelhart, Geoffrey~E Hinton, and Ronald~J Williams,
\newblock ``Learning representations by back-propagating errors,''
\newblock {\em nature}, vol. 323, no. 6088, pp. 533--536, 1986.

\bibitem{jaderberg2017decoupled}
Max Jaderberg, Wojciech~Marian Czarnecki, Simon Osindero, Oriol Vinyals, Alex
  Graves, David Silver, and Koray Kavukcuoglu,
\newblock ``Decoupled neural interfaces using synthetic gradients,''
\newblock in {\em International conference on machine learning}. PMLR, 2017,
  pp. 1627--1635.

\bibitem{xie2003equivalence}
Xiaohui Xie and H~Sebastian Seung,
\newblock ``Equivalence of backpropagation and contrastive hebbian learning in
  a layered network,''
\newblock {\em Neural computation}, vol. 15, no. 2, pp. 441--454, 2003.

\bibitem{bengio2015early}
Yoshua Bengio and Asja Fischer,
\newblock ``Early inference in energy-based models approximates
  back-propagation,''
\newblock {\em arXiv preprint arXiv:1510.02777}, 2015.

\bibitem{scellier2017equilibrium}
Benjamin Scellier and Yoshua Bengio,
\newblock ``Equilibrium propagation: Bridging the gap between energy-based
  models and backpropagation,''
\newblock {\em Frontiers in computational neuroscience}, vol. 11, pp. 24, 2017.

\bibitem{nokland2016direct}
Arild N{\o}kland,
\newblock ``Direct feedback alignment provides learning in deep neural
  networks,''
\newblock {\em Advances in neural information processing systems}, vol. 29,
  2016.

\bibitem{neftci2017event}
Emre~O Neftci, Charles Augustine, Somnath Paul, and Georgios Detorakis,
\newblock ``Event-driven random back-propagation: Enabling neuromorphic deep
  learning machines,''
\newblock {\em Frontiers in neuroscience}, vol. 11, pp. 324, 2017.

\bibitem{belilovsky2020decoupled}
Eugene Belilovsky, Michael Eickenberg, and Edouard Oyallon,
\newblock ``Decoupled greedy learning of cnns,''
\newblock in {\em International Conference on Machine Learning}. PMLR, 2020,
  pp. 736--745.

\bibitem{lowe2019putting}
Sindy L{\"o}we, Peter O'Connor, and Bastiaan Veeling,
\newblock ``Putting an end to end-to-end: Gradient-isolated learning of
  representations,''
\newblock {\em Advances in neural information processing systems}, vol. 32,
  2019.

\bibitem{pyeon2020sedona}
Myeongjang Pyeon, Jihwan Moon, Taeyoung Hahn, and Gunhee Kim,
\newblock ``Sedona: Search for decoupled neural networks toward greedy
  block-wise learning,''
\newblock in {\em International Conference on Learning Representations}, 2020.

\bibitem{he2016deep}
Kaiming He, Xiangyu Zhang, Shaoqing Ren, and Jian Sun,
\newblock ``Deep residual learning for image recognition,''
\newblock in {\em Proceedings of the IEEE conference on computer vision and
  pattern recognition}, 2016, pp. 770--778.

\bibitem{simonyan2014very}
Karen Simonyan and Andrew Zisserman,
\newblock ``Very deep convolutional networks for large-scale image
  recognition,''
\newblock {\em arXiv preprint arXiv:1409.1556}, 2014.

\bibitem{nokland2019training}
Arild N{\o}kland and Lars~Hiller Eidnes,
\newblock ``Training neural networks with local error signals,''
\newblock in {\em International conference on machine learning}. PMLR, 2019,
  pp. 4839--4850.

\bibitem{caporale2008spike}
Natalia Caporale and Yang Dan,
\newblock ``Spike timing--dependent plasticity: a hebbian learning rule,''
\newblock {\em Annu. Rev. Neurosci.}, vol. 31, pp. 25--46, 2008.

\bibitem{yin2023anatomically}
Chenzhong Yin, Phoebe Imms, Mingxi Cheng, Anar Amgalan, Nahian~F Chowdhury,
  Roy~J Massett, Nikhil~N Chaudhari, Xinghe Chen, Paul~M Thompson, Paul Bogdan,
  et~al.,
\newblock ``Anatomically interpretable deep learning of brain age captures
  domain-specific cognitive impairment,''
\newblock {\em Proceedings of the National Academy of Sciences}, vol. 120, no.
  2, pp. e2214634120, 2023.

\bibitem{imms2023early}
Phoebe Imms, Jaron Kawamura, Chenzhong Yin, Anar Amgalan, Nahian~F Chowdhury,
  Nikhil~N Chaudhari, and Andrei Irimia,
\newblock ``Early identification of alzheimer’s risk using anatomically
  interpretable estimation of brain age,''
\newblock in {\em Alzheimer's Association International Conference}. ALZ, 2023.

\bibitem{szegedy2015going}
Christian Szegedy, Wei Liu, Yangqing Jia, Pierre Sermanet, Scott Reed, Dragomir
  Anguelov, Dumitru Erhan, Vincent Vanhoucke, and Andrew Rabinovich,
\newblock ``Going deeper with convolutions,''
\newblock in {\em Proceedings of the IEEE conference on computer vision and
  pattern recognition}, 2015, pp. 1--9.

\bibitem{abadi2016tensorflow}
Mart{\'\i}n Abadi, Paul Barham, Jianmin Chen, Zhifeng Chen, Andy Davis, Jeffrey
  Dean, Matthieu Devin, Sanjay Ghemawat, Geoffrey Irving, Michael Isard,
  et~al.,
\newblock ``Tensorflow: a system for large-scale machine learning.,''
\newblock in {\em Osdi}. Savannah, GA, USA, 2016, vol.~16, pp. 265--283.

\bibitem{krizhevsky2009learning}
Alex Krizhevsky, Geoffrey Hinton, et~al.,
\newblock ``Learning multiple layers of features from tiny images,''
\newblock 2009.

\bibitem{le2015tiny}
Ya~Le and Xuan Yang,
\newblock ``Tiny imagenet visual recognition challenge,''
\newblock {\em CS 231N}, vol. 7, no. 7, pp. 3, 2015.

\end{thebibliography}

\end{document}